\pdfoutput=1
\documentclass[10pt,twocolumn,letterpaper]{article}

\usepackage[pagenumbers]{cvpr} 

\usepackage{graphicx}
\usepackage{amsmath}
\usepackage{amssymb}
\usepackage{booktabs}
\usepackage{bbding}
\usepackage{multirow}
\usepackage[table]{xcolor}
\usepackage{colortbl}
\usepackage[accsupp]{axessibility}  
\usepackage{appendix}

\usepackage[pagebackref,breaklinks,colorlinks]{hyperref}

\usepackage[capitalize]{cleveref}
\crefname{section}{Sec.}{Secs.}
\Crefname{section}{Section}{Sections}
\Crefname{table}{Table}{Tables}
\crefname{table}{Tab.}{Tabs.}

\begin{document}

\title{Dense-Localizing Audio-Visual Events in Untrimmed Videos:\\ A Large-Scale Benchmark and Baseline}

\author{
Tiantian Geng$^{1,2}$, Teng Wang$^{1,3}$, Jinming Duan$^{2}$, Runmin Cong$^{4}$, Feng Zheng$^{1,5*}$ \\
$^1$Southern University of Science and Technology \ $^2$University of Birmingham\\
$^3$The University of Hong Kong \ $^4$Shandong University  \ $^5$Peng Cheng Laboratory \\
{\tt\small gengtiantian97@gmail.com\ tengwang@connect.hku.hk\ j.duan@bham.ac.uk }
\ \ \\ {\tt\small  \ \ rmcong@sdu.edu.cn \ \ f.zheng@ieee.org}
}


\maketitle

\begin{abstract}

\let\thefootnote\relax\footnotetext{{$*$ Corresponding author}}

Existing audio-visual event localization (AVE) handles manually trimmed videos with only a single instance in each of them. 
However, this setting is unrealistic as natural videos often contain numerous audio-visual events with different categories. 
To better adapt to real-life applications, 
in this paper we focus on the task of dense-localizing audio-visual events, which aims to jointly localize and recognize all audio-visual events occurring in an untrimmed video.
The problem is challenging as it requires fine-grained audio-visual scene and context understanding.
To tackle this problem, we introduce the first Untrimmed Audio-Visual (UnAV-100) dataset, 
which contains 10K untrimmed videos with over 30K audio-visual events. 
Each video has 2.8 audio-visual events on average, and the events are usually related to each other and might co-occur as in real-life scenes.
Next, we formulate the task using a new learning-based framework, which is capable of fully integrating audio and visual modalities to localize audio-visual events with various lengths and capture dependencies between them in a single pass.
Extensive experiments demonstrate the effectiveness of our method as well as the significance of multi-scale cross-modal perception and dependency modeling for this task.
\end{abstract}

\section{Introduction}
\label{sec:intro}
Understanding real-world scenes and events is inherently a multisensory perception process for humans~\cite{spence2007audiovisual,  kayser2015multisensory}.
However, for machines, how to integrate multi-modal information, especially audio and visual ones, to facilitate comprehensive video understanding is still a challenging problem. 
In recent years, with the introduction of many audio-visual datasets~\cite{gemmeke2017audio, chen2020vggsound, tian2018audio, Chen_2021_CVPR}, we have seen progress in learning joint audio-visual representations~\cite{arandjelovic2017look, nagrani2021attention,Owens_2018_ECCV}, spatially localizing visible sound sources~\cite{Chen_2021_CVPR, liu2022visual} and temporally localizing audio-visual events~\cite{wu2019dual,zhou2021positive, xia2022cross}, \etc.
\begin{figure}[t]
  \centering
  \setlength{\abovecaptionskip}{1.5mm}
   \includegraphics[width=1.0\linewidth]{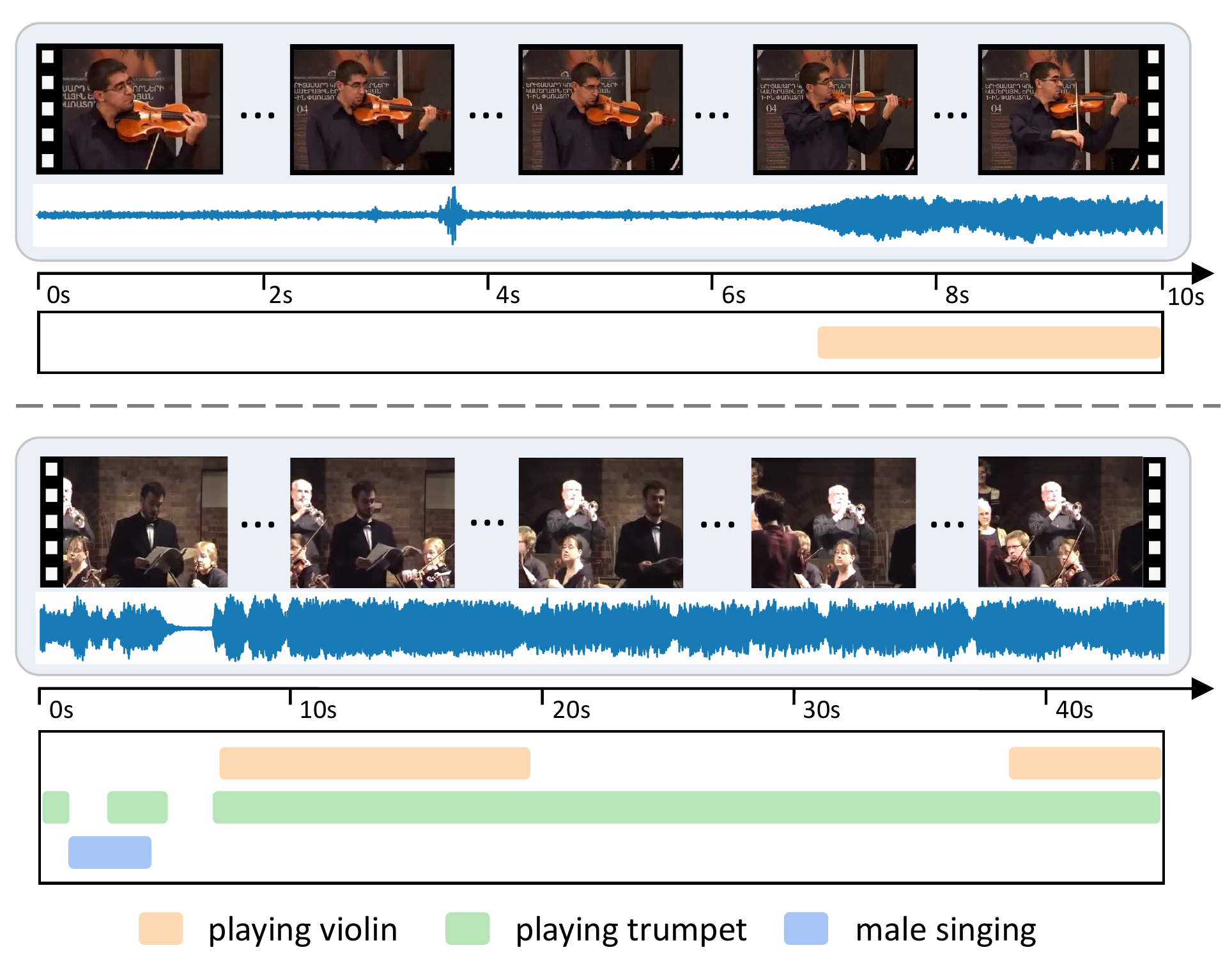}
  \caption{
Different from the previous AVE task, dense-localizing audio-visual events involves localizing and recognizing all audio-visual events occurring in an untrimmed video.
In real-life audio-visual scenes, there are often multiple audio-visual events that might be very short or long, and occur concurrently. The top and bottom examples are from the current AVE dataset~\cite{tian2018audio} and our UnAV-100 dataset, respectively.}
   \label{fig:fig1}
  \vspace{-4.5mm}
\end{figure}
While the success of these algorithms is encouraging, they all focus on manually trimmed videos that often just contain a single audio-visual instance/object in each of them.
In particular, audio-visual event localization (AVE)~\cite{tian2018audio} aims to localize a single event that is both audible and visible at the same time in a short, trimmed video, as shown in the upper part of Fig.~\ref{fig:fig1}.
The task setting is impractical as a real-life video is usually long, untrimmed and associated to multiple audio-visual events from different categories, and these events might have various duration and occur simultaneously.
For example, as illustrated at the bottom of Fig.~\ref{fig:fig1}, a man starts singing and other people accompany him on trumpet and violin, and they pause several times along with the music.
Therefore, we argue that it is necessary to re-examine and re-define the AVE task to better adapt to real-life audio-visual scenarios. 

In this work, we conduct in-depth research starting from dataset construction to technical solutions.
On the one hand, different from the existing AVE dataset~\cite{tian2018audio} that only contains a single audio-visual event in each 10s trimmed video, we introduce a large-scale \textit{Untrimmed Audio-Visual} (UnAV-100) dataset. It consists of more than 10K untrimmed videos with over 30K audio-visual events covering 100 different event categories. 
Our dataset spans a wide range of domains, including human activities, music performances, and sounds from animals, vehicles, tools, nature, \etc.    
As the first audio-visual dataset built on untrimmed videos, UnAV-100 is quite challenging for many reasons.
For instance, each video contains 2.8 audio-visual events on average (23 events maximum), and around $25\%$ of videos have concurrent events.
Besides, the length of audio-visual events varies greatly from 0.2s to 60s. 
There are also rich temporal dependencies among events occurring in a video, \eg., people often clap when cheering, and rain is usually with thunder, \etc.
We believe that the UnAV-100 dataset, with its realistic complexity, can promote the exploration on comprehensive audio-visual video understanding.

On the other hand, facing such a complex real-life scene, current methods\cite{tian2018audio, wu2019dual, xu2020cross, zhou2021positive, xia2022cross} formulate the AVE task as a single-label segment-level classification problem and can only identify one audio-visual event for each segment in a trimmed video. 
They fail to locate concurrent events and provide an exact temporal extent for each event in untrimmed videos. 
To address the above issues, we re-define AVE as an instance-level localization problem, called \textit{dense-localizing audio-visual events}. 
We also present a new framework to flexibly recognize all audio-visual events in an untrimmed video and meanwhile regress their temporal boundaries in a single pass. 
Firstly, the sound and its visual information are both critical to identify an audio-visual event, and the events can range across multiple time scales.
Hence, we propose a cross-modal pyramid transformer encoder that enables the model to fully integrate informative audio and visual signals and capture both very short as well as long audio-visual events.
Secondly, with the observation that the events in a video are usually related to one another, we conduct temporal dependency modeling to learn such correlations, allowing the model to use context to localize events more correctly.
Finally, we design a class-aware regression head for decoding temporal boundaries of overlapping events, together with a classification head to obtain the final localization results.
Extensive experiments demonstrate the effectiveness of our method, and show that it outperforms related state-of-the-art methods for untrimmed videos by a large margin.

Our contributions can be summarized as follows:
\vspace{-2mm}
\begin{itemize}
    \item We introduce a large-scale UnAV-100 dataset,  
    as the first audio-visual benchmark based on untrimmed videos. There exist multiple audio-visual events in each video, and these events are usually related to one another and co-occur as in real-life scenes. 
    \vspace{-2mm}
    \item We shift the AVE task to a more realistic setup of \textit{dense-localizing audio-visual events}, and propose a new framework, allowing to flexibly recognize all audio-visual events in an untrimmed video and regress their temporal boundaries in a single pass.  
    \vspace{-2mm}
    \item Extensive experiments demonstrate the significance of multi-scale cross-modal perception and dependency modeling for the task. Our method can achieve superior performance over related state-of-the-art methods for untrimmed videos by a large margin.
\end{itemize}

\section{Related Work} 
\subsection{Uni-Modal Temporal Localization Tasks}
Deep learning methods have achieved promising performance in temporally localizing target instances using one modality as input, including temporal action localization (TAL) and sound event detection (SED) tasks.
{\bf Temporal action localization (TAL)} aims to detect and classify actions in untrimmed videos. It can be divided into two-stage and single-stage approaches.
A two-stage TAL approach first generates action boundaries with confidence scores, and then classifies their corresponding segments into action categories and refines the generated temporal boundaries~\cite{lin2019bmn,lin2018bsn,bai2020boundary,xu2020g}. 
By contrast, single-stage TAL localizes actions in a single shot without using pre-generated proposals, including anchor-based~\cite{long2019gaussian} and anchor-free methods~\cite{lin2021learning,yang2020revisiting}. 
Besides, Transformers~\cite{vaswani2017attention}, with its powerful ability of long-range relation modeling, are recently also considered in some single-stage TAL methods~\cite{tan2021relaxed,zhang2022actionformer,liu2022end}.
{\bf Sound event detection (SED)} focuses on recognizing and locating audio events in pure acoustic environments~\cite{mesaros2016metrics}. Approaches~\cite{mesaros2016tut,cakir2015polyphonic,parascandolo2016recurrent} cast it as a classification problem to classify the sound category for each temporal unit. 
Overall, both of them belong to uni-modal temporal localization tasks, \ie, TAL detects visual actions, ignoring the auditory information, while SED only considers sound tracks without utilizing visual content. 
Thus, they are both not beneficial for joint audio-visual scene understanding.  

\subsection{Audio-Visual Event Localization}
Tian \etal~\cite{tian2018audio} first proposed the audio-visual event localization task and introduced the AVE dataset. 
Afterward, Wu \etal~\cite{wu2019dual} presented a dual attention matching module for better high-level event information modeling and also attaining local temporal cues. 
Xu \etal~\cite{xu2020cross} designed a relation-aware module to build connections between visual and audio modalities. 
Besides, a positive sample propagation module was proposed by Zhou \etal~\cite{zhou2021positive} to adaptively aggregate positive audio-visual pairs and avoid interference of irrelevant pairs. 
Yan \etal~\cite{xia2022cross} devised a background suppression scheme to suppress cross-modal asynchronous information and uni-modal noise. 
However, all these methods treat the task as a single-label segment classification problem, which fails to localize multiple, concurrent events in an untrimmed video.
In addition, AVE~\cite{tian2018audio} dataset is based on manually trimmed, short videos that only contain a single audio-visual event in each of them, which is inconsistent with real-life audio-visual scenes.
On the other hand, the recent audio-visual video parsing task~\cite{tian2020unified} aims to identify multiple audio, visual and audio-visual events occurring in videos. However, the methods~\cite{tian2020unified,zhou2021positive,lin2021exploring} are also based on simple, trimmed videos in LLP dataset~\cite{tian2020unified}, and can be only deployed in a weekly-supervised manner.

\section{The UnAV-100 Dataset}
\subsection{Overview}
To explore audio-visual event localization in more practical scenes, we build a large-scale UnAV-100 dataset, as the first audio-visual dataset for untrimmed videos.
Each video usually contains multiple audio-visual events annotated with categories and accurate temporal boundaries. The events can be very short as well as long and even overlap in time.
Besides, the dataset covers a wide range of domains, including human activities, music performances, animals/vehicles/tools/natural sounds, \etc.

\begin{table}
  \centering
  \resizebox{\linewidth}{!}{ 
  \begin{tabular}{l|ccccccc }
    \toprule
    Dataset & Videos & Classes & Avg. Length & Annotations & TB & ME  \\
    \midrule
    \midrule
    AudioSet~\cite{gemmeke2017audio} & 2.1M & 527 & 10s & A & \XSolidBrush & \XSolidBrush \\
    Kinetics-Sound~\cite{arandjelovic2017look} & 19K & 34 & 10s & V & \XSolidBrush & \XSolidBrush \\
    VGGSound~\cite{chen2020vggsound} & 200K & 300 & 10s & AV & \XSolidBrush & \XSolidBrush  \\
    ACAV100M~\cite{lee2021acav100m} & 100M & - & 10s & weak AV & \XSolidBrush &\XSolidBrush \\
    \midrule
    AVE~\cite{tian2018audio} & 4,143 & 28 & 10s & AV & \Checkmark & \XSolidBrush  \\
    LLP~\cite{tian2020unified} & 11,849 & 25 & 10s & weak A, V & \Checkmark & \Checkmark \\
    \midrule
    UnAV-100 (Ours) & 10,790 & 100 & 42.1s & AV & \Checkmark &  \Checkmark \\
    \bottomrule
  \end{tabular}}
  \caption{Comparison with related audio-visual datasets. A: audio events; V: visual events; AV: audio-visual events; TB: temporal boundaries; ME: multiple events.}
  \label{tab:comparison}
  \vspace{-4mm}
\end{table}
The comparison with other related audio-visual datasets is shown in Tab.~\ref{tab:comparison}. 
The datasets in the top rows are mainly designed for audio-visual representation learning. 
They all consist of 10s short clips, and there is only a single video-level label provided for each clip.
Among them, AudioSet~\cite{gemmeke2017audio} annotates videos only based on their sound without considering visual information. 
Kinetics-Sound~\cite{arandjelovic2017look}, as a subset of Kinetics~\cite{kay2017kinetics} for action recognition, is annotated based on visual actions, resulting that many videos contain sound tracks unrelated to the visual content (\eg, background music, offscreen voice).
Besides, the videos in VGGSound~\cite{chen2020vggsound} and ACAV100M~\cite{lee2021acav100m} have relatively good audio-visual correspondence, while they are curated using automatic algorithms leading to numerous noisy data.
And ACAV100M~\cite{lee2021acav100m} just provides weak labels obtained from pre-trained uni-modal classifiers.
AVE~\cite{tian2018audio}, the existing dataset for audio-visual event localization, just contains 4K samples with limited 28 event classes. 
Each video is a trimmed 10s clip containing only one audio-visual event, and most events span over the entire video, which is seriously inconsistent with real-world scenarios. 
LLP~\cite{tian2020unified} is designed for weakly-supervised audio-visual video parsing, only providing video-level weak labels for all training data.
By contrast, our UnAV-100 dataset is based on untrimmed videos,
containing over 10K samples with 100 audio-visual event categories. 
Moreover, there are usually multiple audio-visual events annotated in a video with their categories and precise temporal boundaries.
In the following, we provide detailed descriptions about the dataset construction and statistical analysis of our UnAV-100 dataset.

\subsection{Dataset Construction}
\noindent{\bf Collection.} We select VGGSound~\cite{chen2020vggsound} as our data collection source for its relatively high audio-visual correspondence in videos. 
Specifically, we first chose the categories that are common in our daily life from 300 classes in VGGSound covering diverse domains. 
Then, we downloaded raw videos rather than 10s trimmed clips using the provided YouTube URLs.
Since the lengths of raw videos usually span several hours, we randomly cut them within one minute to ensure reasonable duration, meanwhile keeping the videos containing the original 10s clips.
Afterward, we manually verified the presence of audio-visual events in each obtained video.
We found that, since VGGSound was collected in an automated manner, there exist numerous videos that do not contain any audio-visual events. 
For instance, some videos have correct visual content with unrelated sounds like background music and narrations. 
Additionally, it also contains low-quality and animated videos with unrecognizable events.
Finally, by filtering the above cases, we selected around 10K from downloaded 15K videos for annotation.

\begin{figure*}[ht]
  \centering
  \setlength{\abovecaptionskip}{1.0mm}
   \includegraphics[width=1.0\linewidth]{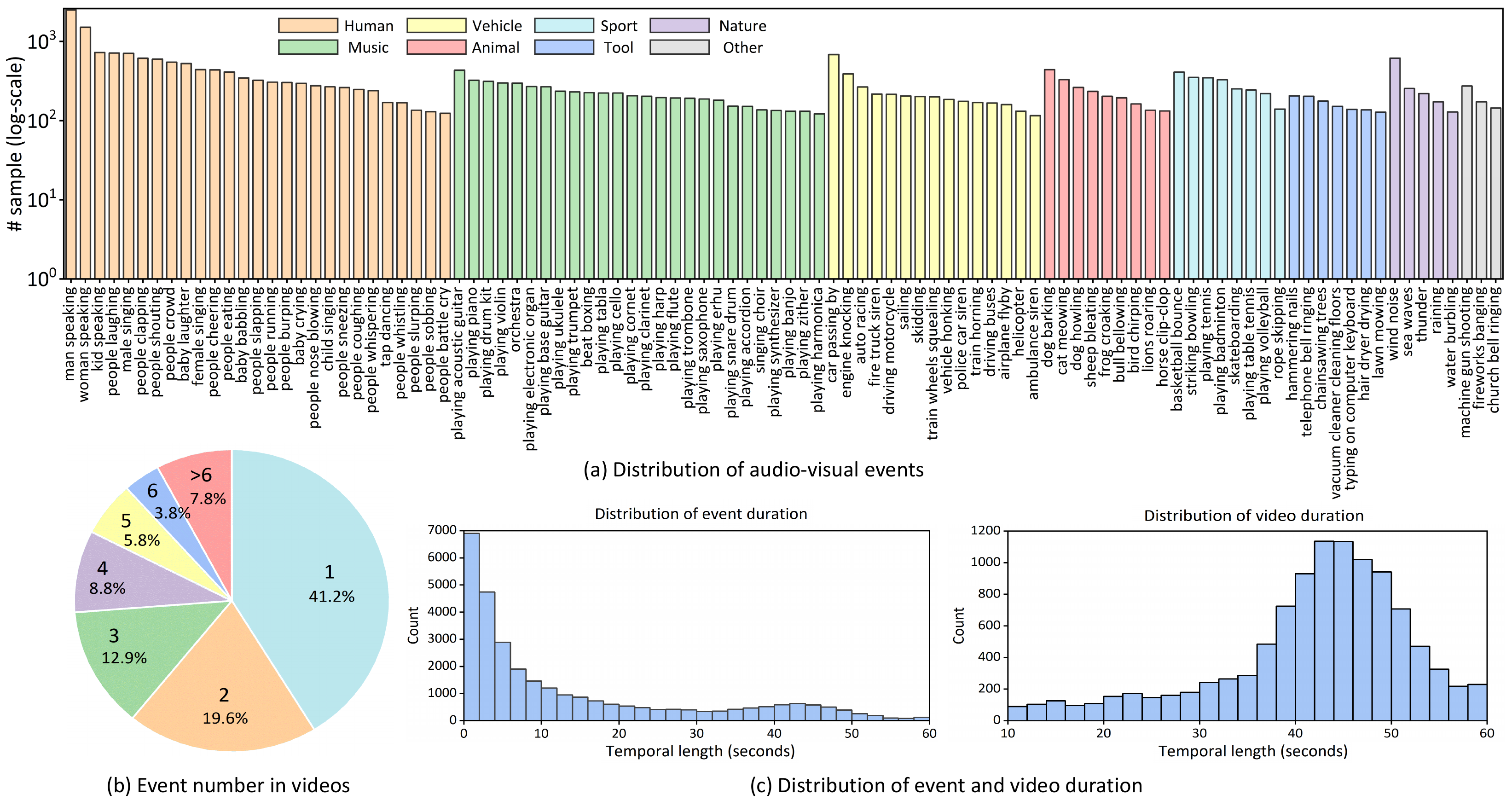}
   \caption{Illustrations of statistics on our UnAV-100 dataset. (a) Distribution of audio-visual events. Bars are grouped by domains, and different colors mean that event categories belong to different domains. 
   (b) The number of audio-visual events in videos. (c) Distribution of event (left) and video duration (right).}
   \label{fig:dataset_overall}
     \vspace{-4mm}
\end{figure*}
\noindent{\bf Annotation.} We annotated videos via an open-source annotation tool VIA~\cite{dutta2019vgg} by crowdsourcing. 
Specifically, we provided expert annotators with a category list for reference, and all audio-visual events occurring in videos are required to be annotated with their categories and independent start and end timestamps. 
Different from the temporal boundaries of visual contents that are usually ambiguous~\cite{caba2015activitynet}, the start and end time points of an audio-visual event are usually clearer and can be easily identified by judging if the event occurs in both audio and visual channels. 
Thus, there is usually high agreement among annotators in labeling temporal boundaries.
In order to ensure annotation quality high, the labeling team is required to check all annotated data carefully. We also employed another group of crowdworkers to manually check it again, resulting in a very time-consuming process.

\subsection{Statistical Analysis}
Overall, our UnAV-100 dataset contains 30,059 audio-visual events of 100 categories, distributed in 10,790 untrimmed videos for over 126 video hours. 
The dataset is split into training, validation, and testing sets with a ratio of 3:1:1,
where a multi-label split strategy~\cite{2017arXiv170201460S} is applied to ensure a well-balanced data distribution in subsets.
Besides, in order to alleviate the effect of long tails, we make sure that there are more than 116 audio-visual events for each category. 
Fig.~\ref{fig:dataset_overall} provides the statistics of our dataset, and the challenges of UnAV-100 include the following:

\noindent{\bf1) Multiple events in videos.} As shown in Fig.~\ref{fig:dataset_overall}(b), around $60\%$ of videos contain more than one audio-visual event.
Each video has 2.8 audio-visual events on average (1.6 for distinct ones), and the maximum number is 23.
Besides, about $25\%$ of videos have concurrent events (the details are in the \textit{Supp. Materials}), which means that there is more than one visible sound source at the same time. 
\begin{figure}[t]
  \centering
  \setlength{\abovecaptionskip}{1.0mm}
   \includegraphics[width=0.9\linewidth]{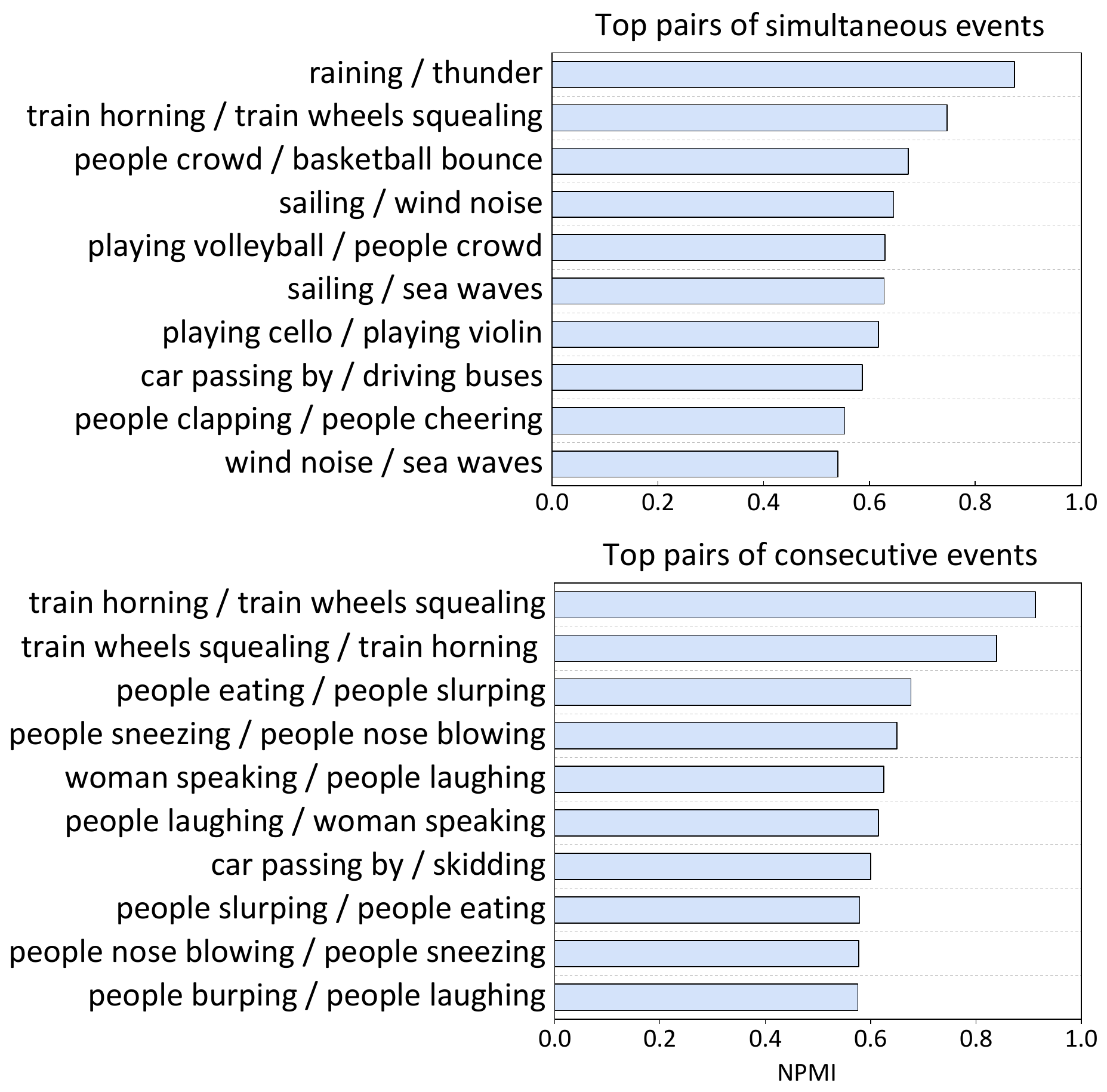}
   \caption{Top pairs of simultaneous and consecutive audio-visual events computed by NPMI falling in the range (-1, 1].}
   \label{fig:NPMI}
   \vspace{-6mm}
\end{figure}

\noindent{\bf2) Various lengths of events and videos.} Fig.~\ref{fig:dataset_overall}(c) shows that a large number of events have very short duration, with the shortest being only 0.2s.
Short events are often difficult to detect, but it aligns with real-life scenes. For example, \textit{dog barking, basketball bounce}, and \textit{fireworks banging} are normally very short audio-visual events.
Besides, the average lengths of audio-visual events and videos are 13.9s and 42.1s, respectively. 
\begin{figure*}[ht]
  \centering
  \setlength{\abovecaptionskip}{0.5mm}
   \includegraphics[width=1.0\linewidth]{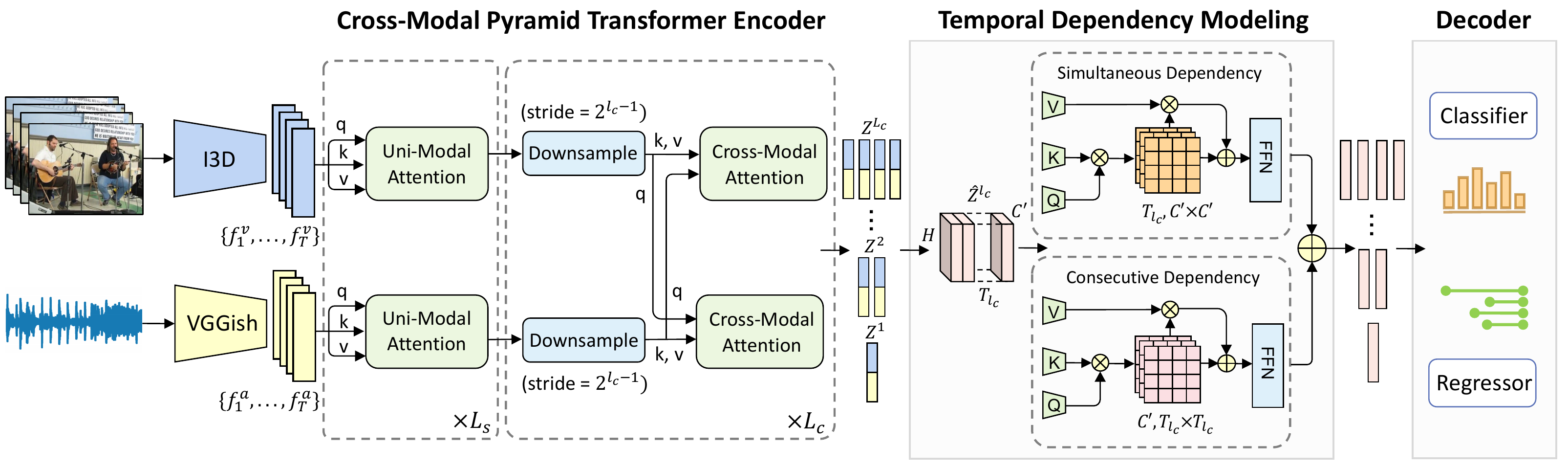}
   \caption{The overview of our proposed framework for dense-localizing audio-visual events. The model takes pre-trained CNNs to extract audio and visual features, and uses a cross-modal pyramid transformer encoder to encode and fuse cross-modal features at various temporal scales, which consists of 
   $L_{s}$ uni-modal and $L_{c}$ cross-modal transformer blocks. 
   Then, the temporal dependency modeling is conducted to capture correlations between events. Finally, a classification and a regression head are used to predict the  categories and temporal boundaries of events in an end-to-end manner.     
   \label{fig:framwork}
   \vspace{-3mm}}
\end{figure*}

\noindent{\bf3) Rich temporal dependencies between events.} The related audio-visual events usually occur simultaneously or consecutively in a video.
In Fig.~\ref{fig:NPMI}, we show the pairs of simultaneous and consecutive events with the top 10 Normalized Pointwise Mutual Information (NPMI)~\cite{church1990word}, respectively. 
We can see, frequently, rain is accompanied by thunder, violin and cello are played together, people clap when cheering, \etc.
Such event dependencies are very similar to real-world intuition and reflect human behavior. 
We note that UnAV-100 is the first audio-visual dataset with such context information,
which provides excellent data for building many complex models for audio-visual event dependency modeling.

\section{Method}
To solve the problem of dense-localizing audio-visual events,
we design an architecture to jointly recognize and localize multiple, concurrent audio-visual events with various lengths, and meanwhile capture event dependencies in an untrimmed video.
An overview of the proposed framework is illustrated in Fig.~\ref{fig:framwork}.

\subsection{Preliminaries}
\noindent{\bf Problem Statement.} Different from previous AVE methods, we formulate the task of dense-localizing audio-visual events as a joint classification and regression problem.
Formally, given an input video sequence containing both visual and audio tracks, we first divide it into $T$ visual and audio segment pairs $\{V_{t}, A_{t}\}_{t=1}^{T}$, where $T$ varies across videos. 
The groundtruth event set for each video is denoted as $Y=\{y_{n}=(t_{s,n}, t_{e,n}, c_{n})\}^{N}_{n=1}$, where $t_{s,n}$, $t_{e,n}$ are the start and end timestamp of $n$-th event, $c_{n}\in\{1, \cdots, C\}$ is the event category, and $N$ is the total number of audio-visual events in the video.
Then, the model is required to predict $\hat{Y}=\{\hat{y}_{t}=(d_{s,t}, d_{e,t}, p(c_{t}))\}^{T}_{t=1}$ during inference, where $p(c_{t})\in \mathbb{R}^{1\times C}$ is the probabilities of $C$ event categories at moment $t$, $d_{s,t}$ and $d_{e, t}$ are the distances between the moment $t$ to the event's start and end timestamp. Note that $d_{s,t}$ and $d_{e,t}$ are only defined when an event presents at moment $t$.
Thus, the final localization results can be obtained by: 
\vspace{-2mm}
\begin{equation}
    c_{t} = {\arg\max}\, p(c_{t}),\quad t_{s,t}=t-d_{s,t},\quad t_{e,t}=t+d_{e,t}.
\label{equ1} 
\vspace{-2mm}
\end{equation}
\noindent{\bf Audio and Visual Representations. }We extract audio feature vectors using the VGGish model~\cite{hershey2017cnn} pre-trained on AudioSet~\cite{gemmeke2017audio}. 
And visual feature vectors are extracted by the two-stream I3D~\cite{carreira2017quo} pre-trained on Kinetics-400~\cite{kay2017kinetics}. 
Then, we apply two convolutional layers with ReLU to project features from two modalities into a shared embedding space,
resulting $F_{V} = \{f^{v}_{t}\}^{T}_{t=1}$, $F_{A}=\{f^{a}_{t}\}^{T}_{t=1} \in \mathbb{R}^{T\times D}$, where $D$ is the dimension of the embedding space.

\subsection{Architecture}
\noindent{\bf Cross-Modal Pyramid Transformer Encoder.} 
We consider that the sound and its corresponding visual information are both crucial to identify an audio-visual event. 
However, the audio and visual tracks of an untrimmed video often contain a lot of irrelevant information (\eg, background music and off-screen voice), and their content might be misaligned with each other (\eg, a dog appears without barking).  
Besides, the events occurring in untrimmed videos usually range across multiple time scales.
Thus, how to appropriately integrate the two modalities and capture very short as well as long events are both significant for this task.
Here, a cross-modal pyramid transformer encoder is proposed to address the above challenges.  

Specifically, in order to capture long-term temporal relations among uni-modal segments and filter out noise in each modality, the feature sequences from two modalities are first fed into $L_{s}$ stacked uni-modal transformer blocks separately. 
Each block regularly contains a multiheaded self-attention (MSA) and a feed-forward network (FFN) with LayerNorm (LN) and residual connections. 
And position embeddings $E_{pos}\in \mathbb{R}^{T \times D}$ as in ~\cite{vaswani2017attention} are also added in input sequences.
By doing this, the model can focus more on event-related information in each modality.
Afterward, the obtained feature sequences are further encoded into a cross-modal pyramid transformer to integrate informative signals from two modalities at different temporal resolutions.
The module consists of $L_{c}$ stacked blocks. 
In each block, as shown in Fig.~\ref{fig:framwork}, we first temporally downsample the feature sequence of each modality with the stride $2^{l_{c}-1}$, where $l_{c}$ is the index of the current block, and the longer strides are able to capture longer events.
Then, we assign downsampled features in the current modality as the key and value vectors, and the features of another modality as the query vector in multiheaded cross-attention (MCA), followed by FFN and LN layers.  
Thus, the audio-guided visual feature $F_{Va}$ and visual-guided audio feature $F_{Av}$ from $l_{c}$-th block can be denoted as:
\begin{equation}
\begin{split}
    &F^{l_{c}}_{Va} = \mathrm{MCA}(\hat{F}^{l_{c}-1}_{Av}W_{q}, \hat{F}^{l_{c}-1}_{Va}W_{k}, \hat{F}^{l_{c}-1}_{Va}W_{v}),\\
    &F^{l_{c}}_{Av} = \mathrm{MCA}(\hat{F}^{l_{c}-1}_{Va}W_{q}, \hat{F}^{l_{c}-1}_{Av}W_{k}, \hat{F}^{l_{c}-1}_{Av}W_{v}),
\end{split}
\end{equation}
where $l_{c}=\{1, \cdots, L_{c}\}$, $F^{l_{c}}_{Va}, F^{l_{c}}_{Av} \in \mathbb{R}^{T_{l_{c}}\times D}$ ($ T_{l_{c}}=T /2^{l_{c}-1}$), 
$\hat{F}^{l_{c}-1}_{Va}$ and $\hat{F}^{l_{c}-1}_{Av}$ are the features after downsampling, 
$W_{q}, W_{k}, W_{v} \in \mathbb{R}^{D\times D_{m}} $ are learnable parameters 
and $D_{m}=D$ is the dimension of learned query, key and value vectors.
After cross-modal interactions at various temporal scales, 
we concatenate the enhanced audio and visual features at the same pyramid level,   
getting a cross-modal feature pyramid $Z=\{Z^{l_{c}}\}^{L_{c}}_{l_{c}=1}$, where $Z^{l_{c}}=\mathrm{Concat}(F^{l_{c}}_{Va},F^{l_{c}}_{Av}) \in \mathbb{R}^{T_{l_{c}}\times 2D}$.

\noindent{\bf Temporal Dependency Modeling.}
The key characteristic of real-life audio-visual scenes is that the related events usually occur simultaneously or consecutively. For example, people are used to clapping when cheering, and cars often honk when passing by. 
Here, inspired by the method~\cite{tirupattur2021modeling} for action dependency modeling in the TAL task, we implicitly capture such simultaneous and consecutive dependencies among audio-visual events at the obtained cross-modal feature pyramid. 
Concretely, for each cross-modal feature sequence $Z^{l_{c}}$, we first transform and expand the feature dimension to $\hat{Z}^{l_{c}}\in \mathbb{R}^{T_{l_{c}}\times C'\times H}$, splitting it into $C'$ groups, where $C'$ represents the number of hidden classes and $H$ is the transformed feature dimension.
We suppose each hidden class is learned to carry a group of distinctive features for event classification. For simultaneous dependency modeling, the self-attention is performed along the $C'$ dimension of $\hat{Z}^{l_{c}}$, which means a $C'\times C'$ attention matrix that denotes the relevance among hidden classes at each time step can be obtained.
For consecutive dependency modeling, the self-attention is performed along $T_{l_{c}}$ dimension, getting a $T_{l_{c}}\times T_{l_{c}}$ attention matrix to indicate the correlations among all time steps for the classification of the given class.
Then the output of the two branches followed by FFN and LN layers with residual connections 
are simply merged by element-wise summation to enable the model to capture both types of dependencies. 
Note that we share the parameters of dependency modeling across all pyramid levels. 

\noindent{\bf Decoder.} Next, a decoder, consisting of a classification head and a regression head, is applied to decode the enhanced feature pyramid into prediction results in a single pass.
Specifically, the classification head predicts the probability $p(c_{t})$ of events at every moment $t$ of all pyramid levels. 
It consists of three layers of 1D convolutions following a sigmoid function as in~\cite{zhang2022actionformer}.
Besides, the regression head outputs the distances to the start and end timestamp of an event $(d_{s,t}, d_{e,t})$ at time step $t$ if the event exists.
We highlight that the regression head is designed to be class-aware, which allows the model to regress temporal boundaries for the overlapping events with different categories.
It is realized by using three 1D convolutions attached with a ReLU, getting the output with the shape of $[2, C, T_{l_{c}}]$ for each pyramid level.
Here, the pyramid architecture enables the regression head to predict temporal boundaries at different temporal scales, allowing the model to capture the events with various lengths. 
Note that the parameters of both two heads are shared across all pyramid levels.

\subsection {Training and Inference}
\noindent{\bf Loss Function.} 
We use two losses to train our model in an end-to-end manner, \ie, a focal loss~\cite{lin2017focal} $\mathcal{L}_{cls}$ for classification and a generalized IoU loss~\cite{rezatofighi2019generalized} $\mathcal{L}_{reg}$ for distance regression, as in the TAL  method~\cite{zhang2022actionformer}.
For each video, the loss function is denoted as:
\begin{equation}
    \mathcal{L}=\frac{1}{\mathcal{T}}\sum_{t}\mathcal{L}_{cls} + \frac{\lambda}{\mathcal{N}}\sum_{t}\mathbb{I}_{t}\mathcal{L}_{reg},
\end{equation}
where $\mathcal{T}$ is the total segment number of all levels, $\mathbb{I}_{t}$ is an indicator function denoting if a timestamp contains events, $\mathcal{N}$ is the number of positive segments that contain events across all levels.
Here, we weight the contribution of $\mathcal{L}_{reg}$ with $\lambda=1$ by default. 

\noindent{\bf Inference. }
During inference, the outputs of the model are as in Eq.~(\ref{equ1})
for every timestamp $t$ across all levels. 
Then the obtained event candidates are post-processed by a multi-class version of Soft-NMS~\cite{bodla2017soft} to suppress redundant temporal boundaries with high overlaps within the same class. 

\section{Experiments}
\subsection{Experimental Settings}
\noindent{\bf Implementation Details.} For each video, we sample frames at 25 fps, and feed 24 consecutive RGB and optical flow frames into two-stream I3D~\cite{carreira2017quo}, using a sliding window with stride 8.
Then, the two-stream features are concatenated (2048-d) as a visual segment. 
Here, the optical flow is extracted by RAFT~\cite{teed2020raft}.
Besides, we extract 128-d audio features by VGGish~\cite{hershey2017cnn} 
for each 0.96s segment with a sliding window (stride=0.32s) to temporally align with the visual ones.
Since the input sequences vary in length, we pad or crop them to the maximum length $T=224$, and add masks for all operations in the model.
The dimensions of the embedding space in the encoder and temporal dependency modeling are $D=512$ and $H=128$, respectively. The number of hidden classes $C'=100$.
Our model is trained with the Adam optimizer, and the number of epochs is 40 with a linear warmup of 5 epochs. The initial learning rate is 1e-4 and a cosine learning rate decay is used. The mini-batch size is 16 and the weight decay is 1e-4.

\noindent{\bf Evaluation Metrics.} As a temporal localization task for untrimmed videos, we use mean Average Precision (mAP) to evaluate results.
Specifically, we report mAPs at the tIoU thresholds [0.5:0.1:0.9] and the average mAP at the thresholds [0.1:0.1:0.9].

\noindent{\bf Baseline Models.} 
Since previous AVE and SED methods are limited to localizing a single event on trimmed videos with the same duration and cannot be applied on untrimmed videos, we only compare our model with recent state-of-the-art TAL models, as shown in Tab.~\ref{tab:compare}.
It includes the two-stage model VSGN~\cite{zhao2021video} and single-stage models (TadTR~\cite{liu2022end} and ActionFormer~\cite{zhang2022actionformer}). 
Here, $L_{s}=2$ and $L_{c}=6$ in the pyramid transformer encoder of our model.
Note that all compared approaches use the same input features as ours to keep a fair comparison.
\begin{table}
  \centering
  \resizebox{\linewidth}{!}{
  \begin{tabular}{c|c|cccccc}
    \toprule
    Modality & Method & 0.5 & 0.6 & 0.7 & 0.8 & 0.9 & Avg.  \\
    \midrule
    \midrule
    \multirow{4}{*}{A}&
    VSGN~\cite{zhao2021video} & 18.0 & 14.2 & 10.8 & 8.2 & 5.3 & 17.8 \\
    & TadTR~\cite{liu2022end} & 23.0 & 20.5 & 17.6 & 14.4 & 10.4 & 22.8 \\
    & ActionFormer~\cite{zhang2022actionformer} & 37.7 & 32.8 & 27.3 & 22.5 & 15.6 & 36.0 \\
    & Ours & 39.0 & 34.5 & 29.1 & 23.3 & 12.5 & 37.1 \\
    \midrule
    \multirow{4}*{V}&
    VSGN~\cite{zhao2021video} & 14.8 & 11.5 & 8.5 & 6.0 & 4.1 & 15.5 \\
    & TadTR~\cite{liu2022end} & 23.1 & 20.5 & 17.8 & 15.3 & 12.0 & 23.0 \\
    & ActionFormer~\cite{zhang2022actionformer} & 36.3 & 31.9 & 27.4 & 21.8 & 14.8 & 35.4 \\
    & Ours & 37.3 & 32.6 & 28.3 & 22.9 & 14.7 & 35.9 \\
    \midrule
    \multirow{4}*{A\&V}&
    VSGN~\cite{zhao2021video} & 24.5 & 20.2 & 15.9 & 11.4 & 6.8 & 24.1 \\
    & TadTR~\cite{liu2022end} & 30.4 & 27.1 & 23.3 & 19.4 & 14.3 & 29.4 \\
    & ActionFormer~\cite{zhang2022actionformer} & 43.5 & 39.4 & 33.4 & 27.3 & 17.9  & 42.2  \\
    & \cellcolor{gray!20}{Ours} & \cellcolor{gray!20}{\bf50.6} & \cellcolor{gray!20}{\bf45.8} & \cellcolor{gray!20}{\bf39.8} & \cellcolor{gray!20}{\bf32.4} & \cellcolor{gray!20}{\bf21.1} & \cellcolor{gray!20}{\bf47.8} \\
    \bottomrule
  \end{tabular}}
  \caption{Comparison of the results on the test set of UnAV-100 dataset. A: only audio modality; V: only visual modality; A\&V: both audio and visual modalities.}
  \label{tab:compare}
  \vspace{-4mm}
\end{table}

\subsection{Results and Analysis}
To validate the effectiveness of the proposed model, we compare it with recent TAL methods using different modalities as input, and also conduct extensive ablation studies.

\noindent{\bf Comparison Results.}
As shown in Tab.~\ref{tab:compare}, when using one modality as input, our model variants that only apply self-attention in the encoder outperform all compared TAL methods, where TadTR~\cite{liu2022end} and ActionFormer~\cite{zhang2022actionformer} also use an end-to-end transformer-based architecture.
When using both audio and visual modalities, 
the performance of our model boosts significantly, \eg, $+11.9\%$ and $+10.7\%$ at the average mAP compared with our visual-only and audio-only variants, respectively. These results clearly indicate that both modalities are equally crucial for this task. 
Besides, our model surpasses the compared TAL methods by a large margin, even though they also benefit greatly from multi-modal input. 
Here, we simply concatenate audio and visual features as input of these methods.
\begin{table}
  \centering
  \resizebox{0.85\linewidth}{!}{
  \begin{tabular}{ccc|cccccc}
    \toprule
    $L_{s}$ & $L_{c}$ & TD & 0.5 & 0.6 & 0.7 & 0.8 & 0.9 & Avg. \\
    \midrule
    \midrule
    2 & 0 &  & 36.8 & 29.3 & 21.8 & 13.8 & 4.9 & 35.5\\
    2 & 1 &  & 37.6 & 29.6 & 22.2 & 14.0 & 5.1 & 35.4 \\
    2 & 2 &  & 37.0 & 29.3 & 20.9 & 12.5 & 3.8 & 35.3 \\
    \midrule
    2 & 2 & \checkmark & 41.0 & 33.1 & 25.7 & 18.0 & 8.1 & 39.4 \\
    2 & 4 & \checkmark & 49.8 & 43.0 & 35.4 & 25.5 & 11.2 & 45.0 \\
    \rowcolor{gray!20} 2 & 6 & \checkmark & {\bf50.6} & {\bf45.8} & {\bf39.8} & 32.4 & 21.1 & {\bf47.8} \\
    2 & 7 & \checkmark & 49.1 & 44.8 & 39.5 & 32.4 & {\bf21.8} & 46.8 \\
    \midrule
    0 & 6 & \checkmark & 49.4 & 45.2 & 39.2 & {\bf32.5} & 21.6 & 46.7 \\
    1 & 6 & \checkmark & 49.6 & 45.3 & 39.5 & {\bf32.5} & 21.3 & 47.0 \\
    3 & 6 & \checkmark & 48.8 & 44.4 & 39.2 & 32.2 & {\bf21.8} & 46.4 \\
    \bottomrule
  \end{tabular}}
  \caption{Ablation study on cross-modal fusion strategies and the design of feature pyramid. TD: temporal downsampling.}
  \label{tab:pyramid}
  \vspace{-2mm}
\end{table}
\begin{table}
  \centering
  \resizebox{0.8\linewidth}{!}{
  \begin{tabular}{cc|cccccc}
    \toprule
    DM & CA & 0.5 & 0.6 & 0.7 & 0.8 & 0.9 & Avg. \\
    \midrule
    \midrule
      &  & 48.2 & 42.3 & 35.5 & 28.0 & 18.1 & 45.2 \\
     \checkmark &  & 48.5 & 44.2 & 38.7 & {\bf32.6} & 21.0 & 46.1 \\
      & \checkmark & 48.5 & 43.4 & 36.9 & 29.9 & 20.2 & 45.8 \\
    \midrule
    \rowcolor{gray!20} \checkmark & \checkmark & {\bf50.6} & {\bf45.8} & {\bf39.8} & 32.4 & {\bf21.1} & {\bf47.8} \\
    \bottomrule
  \end{tabular}}
  \caption{Ablation study on dependency modeling (DM) and class-aware regression (CA).}
  \label{tab:cooccur-multilabel}
  \vspace{-2mm}
\end{table}
\begin{table}
  \centering
  \resizebox{1.0\linewidth}{!}{
  \begin{tabular}{l|cccccc}
    \toprule
    Model & 0.5 & 0.6 & 0.7 & 0.8 & 0.9 & Avg. \\
    \midrule
    \midrule
    ResNet50~\cite{he2016deep} (RGB) & 46.6 & 42.2 & 37.2 & 30.8 & 20.1 & 44.3 \\
    I3D~\cite{carreira2017quo} (RGB) & 49.1 & 44.8 & 39.0 & 32.0 & {\bf21.3} & 46.7 \\
    \rowcolor{gray!20} I3D~\cite{carreira2017quo} (RGB + Flow) & {\bf50.6} & {\bf45.8} & {\bf39.8} & {\bf32.4} & 21.1 & {\bf47.8} \\
    \bottomrule
  \end{tabular}}
  \caption{Ablation study on different visual features.}
  \label{tab:motion feature}
  \vspace{-5mm}
\end{table}
\begin{figure*}[ht]
  \centering
  \setlength{\abovecaptionskip}{1.0mm}
   \includegraphics[width=1.0\linewidth]{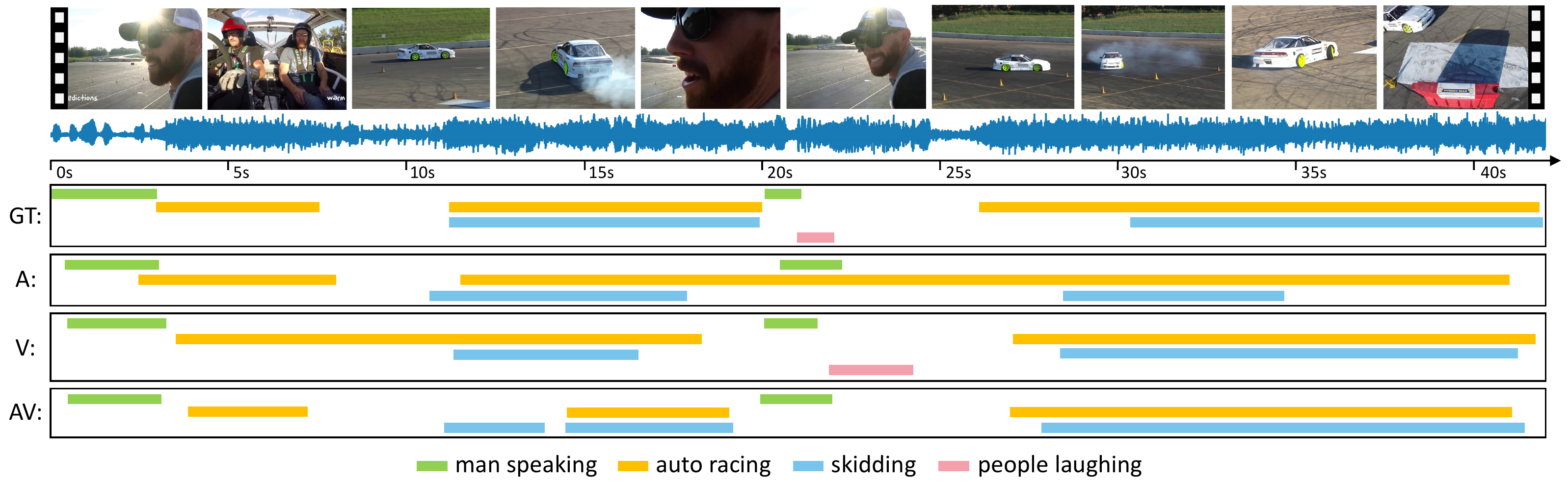}
   \caption{Qualitative results on the UnAV-100 test set. 
   GT: ground truth, A: the prediction of our audio-only variant, V: the prediction of our visual-only variant, AV: the prediction of our audio-visual model. We show boundaries with the highest overlap with ground truth.}
   \label{fig:visualization}
   \vspace{-4mm}
\end{figure*}

\noindent{\bf Cross-Modal Fusion and Pyramid Levels.} We explore the cross-modal fusion strategies and the design of the cross-modal feature pyramid.
In Tab.~\ref{tab:pyramid}, we can see that using only two uni-modal transformer blocks ($L_{s}=2$ and $L_{c}=0$) for each modality separately decreases the performance dramatically.
Later, adding one or two cross-modal blocks at the original temporal resolution can just slightly increase mAP scores.
Instead, applying temporal downsampling in cross-modal blocks boosts the performance, indicating that the cross-modal fusion at multiple temporal resolutions is essential for our model. 
Then, the performance gradually increases by further adding cross-modal pyramid levels, and yet is saturated when $L_{c}=6$.
In addition, we found that the appropriate number of uni-modal blocks is also important,
which reveals that applying self-attention before cross-modal interaction can help the model to focus on informative signals and eliminate noise from each modality.

\noindent{\bf Dependency Modeling and Class-Aware Regression.} 
As shown in Tab.~\ref{tab:cooccur-multilabel}, applying temporal dependency modeling and class-aware regression separately can both achieve higher performances than the base model that just contains our transformer encoder with a class-agnostic regression head in the decoder.
Besides, we found that when using both of them, they can promote each other and achieve a further significant performance boost, which clearly demonstrates their effectiveness.
\begin{figure}[t]
  \centering
  \setlength{\abovecaptionskip}{1.5mm}
   \includegraphics[width=0.6\linewidth]{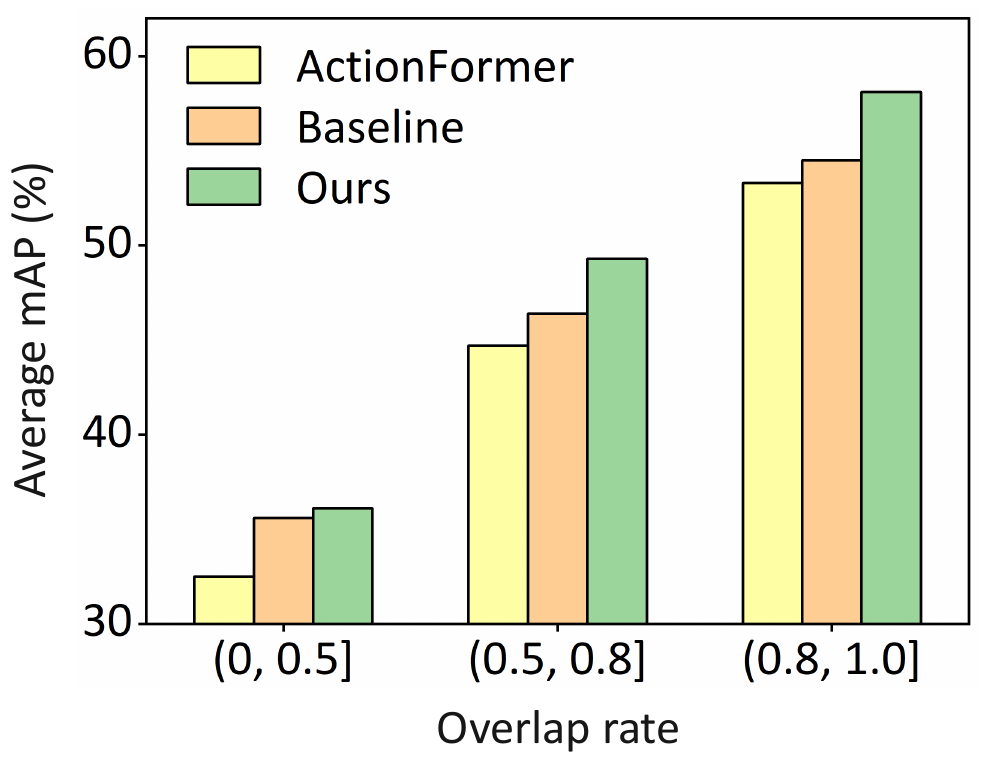}
   \caption{Performance comparison of our models and the TAL method (ActionFormer~\cite{zhang2022actionformer}) on the videos from the UnAV-100 test set, containing concurrent events with different overlap rates.
   }
   \label{fig:overlaprate}
   \vspace{-5mm}
\end{figure}

\noindent{\bf The Impact of Motion Features.}
In Tab.~\ref{tab:motion feature},
we observe that utilizing both RGB and optical flow features extracted by I3D~\cite{carreira2017quo} achieves the best performance.
It outperforms the model that uses visual features extracted by ResNet50~\cite{he2016deep} pre-trained on ImageNet by a large margin ($+3.5\%$ at the average mAP).
Even though it is proved in ~\cite{tian2018audio} that motion features are useless for audio-visual event localization, we argue that our experiment clearly demonstrates their significance for dense-localizing audio-visual events. 

\noindent{\bf The Capability of Localizing Concurrent Events.}
We further evaluate our models and the state-of-the-art TAL method~\cite{zhang2022actionformer} on the videos that contain concurrent events with different overlap rates in Fig.~\ref{fig:overlaprate}. 
We observe that our model equipped with dependency modeling and class-aware regression obviously gains more performance improvement on the videos with higher overlap rates, compared with our baseline and ActionFormer~\cite{zhang2022actionformer}.
It suggests that our model has a better ability to localize overlapping audio-visual events in untrimmed videos.

\noindent{\bf Qualitative Results.} In Fig.~\ref{fig:visualization}, we present the qualitative results of our model variants that utilize different modalities as input.
We observe that the model using both modalities can localize audio-visual events more correctly, even though some events occur simultaneously or have short duration.
By contrast, since the sound of \textit{auto racing} almost spans the whole video, the audio-only model gets the wrong boundaries of the event without the help of visual information. And similar errors also occur when using the visual-only model.
Overall, it demonstrates again that audio and visual modalities complement each other and are equally significant for dense-localizing audio-visual events. 
More ablation studies and qualitative results can be found in the \textit{Supp. Materials}.

\section{Conclusion}
In this work, we investigate the dense-localizing audio-visual events problem, which aims to recognize and localize all audio-visual events occurring in an untrimmed video. 
To facilitate this research, we build a large-scale UnAV-100 dataset consisting of more than 10K untrimmed videos with over 30K audio-visual events covering 100 categories.
We also propose a new framework, formulating the task as a joint classification and regression problem, which is capable of localizing audio-visual events that have various lengths and overlap in time, and capturing the dependencies between them in a video.
Our results demonstrate the superiority of our model, indicating the significance of cross-modal perception and dependency modeling for this task.

\vspace{+2mm}
\noindent{\bf Acknowledgments.} This work was supported by the National Key R\&D Program of China (Grant NO. 2022YFF1202903) and the National Natural Science Foundation of China (Grant NO. 62122035 and 61972188).

{\small
\bibliographystyle{ieee_fullname}
\bibliography{arxiv_paper}
}
\newpage
\section*{Appendix}
\appendix
\section{More Statistical Analysis}
\begin{figure}[t]
  \centering
  \setlength{\abovecaptionskip}{2mm}
   \includegraphics[width=0.8\linewidth]{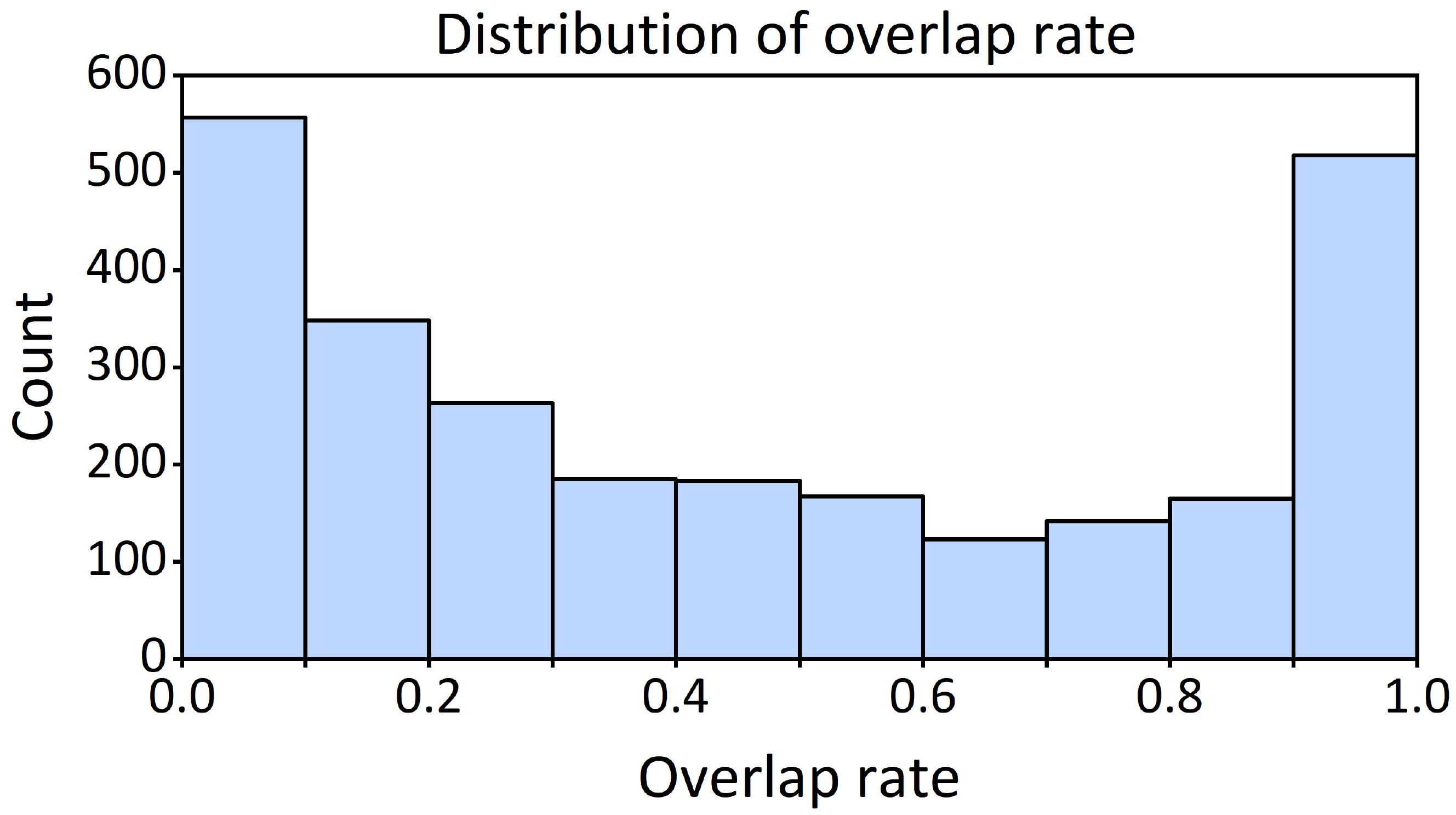}
  \caption{Overlap rate distribution of the videos that contain concurrent events in our UnAV-100 dataset.}
   \label{fig:overlap}
\end{figure}

\noindent{\bf Concurrent Events.} There are usually multiple audio-visual events occurring simultaneously in UnAV-100 dataset as in real-life scenes. 
Here, we define the overlap rate $\mathcal{O}$ of each video as:
\begin{equation}
    \mathcal{O} = \frac{U_{o}}{U_{e}},
\end{equation}
where $U_{o}$ is the temporal union of overlapping intervals, and $U_{e}$ is the temporal union of the intervals of all audio-visual events in the video.
Totally, there are around $25\%$ of videos (2,651) containing concurrent audio-visual events ($\mathcal{O} > 0.01$, considering annotation errors) in our UnAV-100 dataset.
The overlap rate distribution of these videos is illustrated in Fig.~\ref{fig:overlap}.
We can see that the videos with low and high overlap rates both have high proportions.
Higher overlap rates might indicate that the events have higher correlations and usually occur at the same time, which requires the model to have a strong ability of dependency modeling. 

\begin{figure*}[ht]
  \centering
  \setlength{\abovecaptionskip}{2mm}
  \subfloat[\label{1a}]{
   \includegraphics[width=0.72\linewidth]{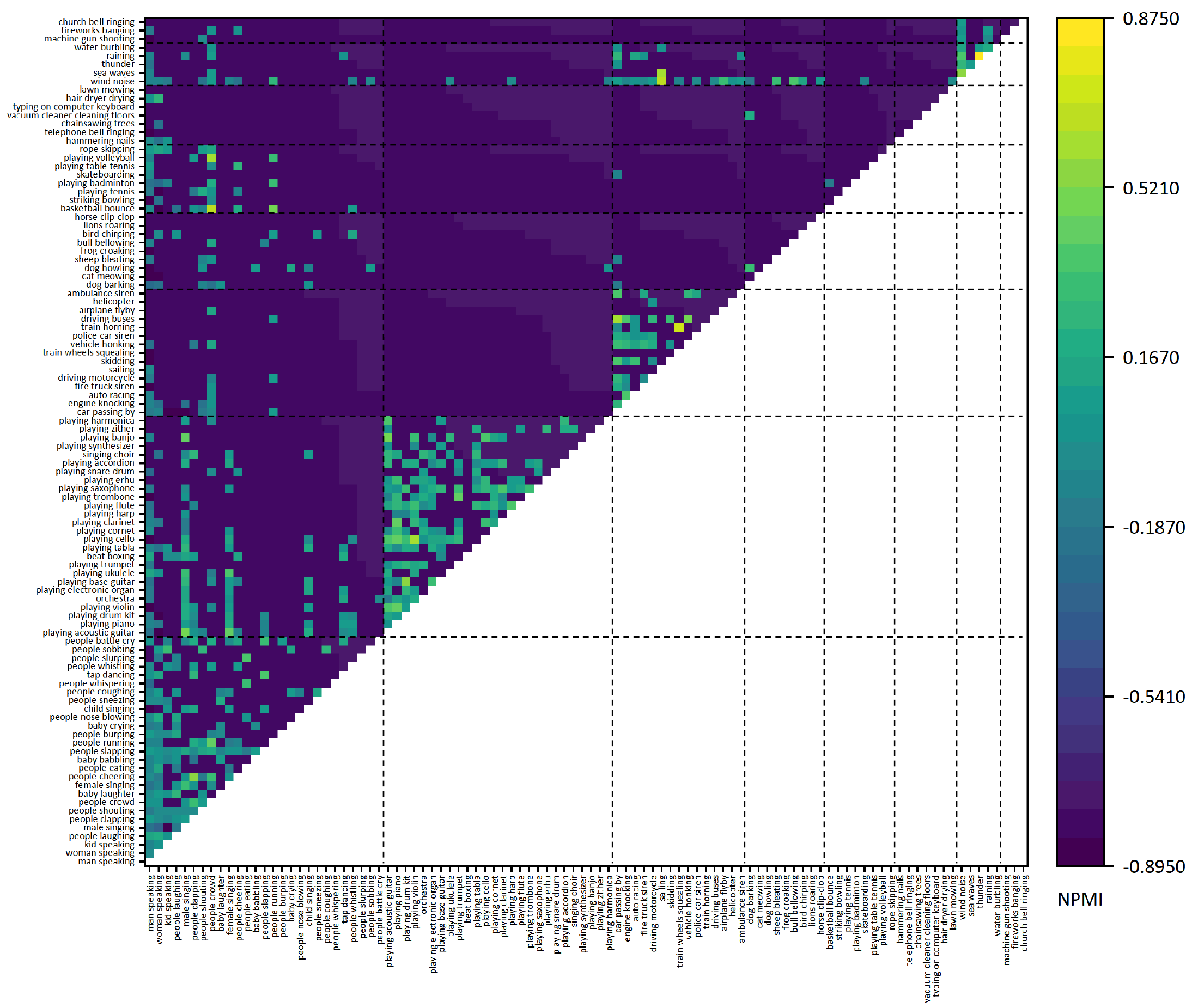}}\\
    \subfloat[\label{1b}]{
   \includegraphics[width=0.72\linewidth]{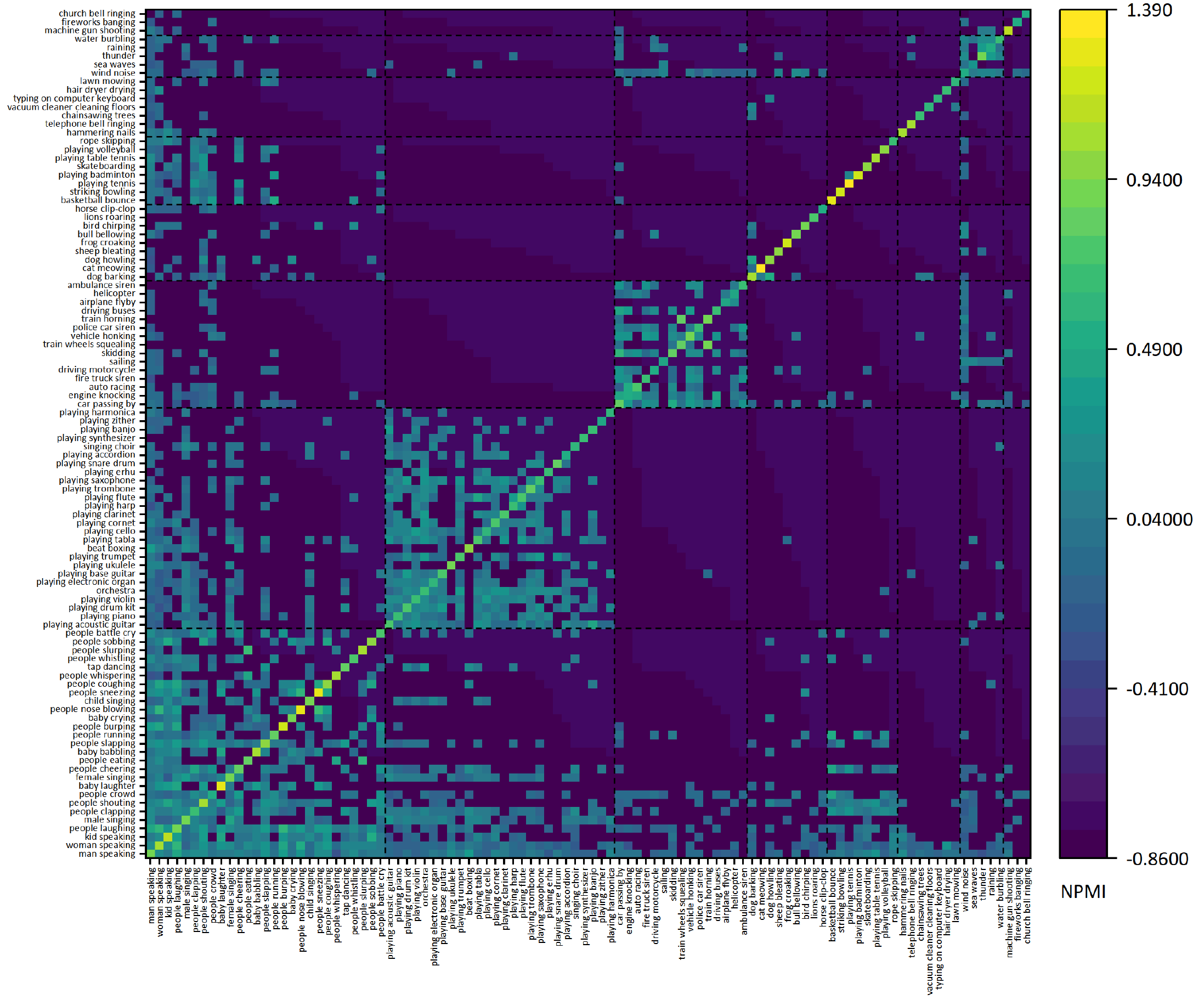}}
  \caption{NPMI of the pairs of simultaneous (a) and consecutive (b) audio-visual events in our UnAV-100 dataset. 
  In (b), the horizontal axis shows the first event, and the vertical axis shows the second subsequent event.
  The event categories are grouped by domains.}
   \label{fig:npmi}
\end{figure*}

\noindent{\bf Temporal Dependencies between Events.} We show NPMI (Normalized Pointwise Mutual Information)~\cite{church1990word} of the pairs of simultaneous and consecutive audio-visual events for all 100 event categories in Fig.~\ref{fig:npmi}(a) and Fig~\ref{fig:npmi}(b), respectively. NPMI is commonly used in linguistics to represent the co-occurrence between two words.
Firstly, in Fig.~\ref{fig:npmi}(a), we can observe that the event categories from the same domains are more likely to occur concurrently, \eg, the events of human activities, music performances, and the sounds of vehicles/natural.
Besides, the events from various domains are usually accompanied by human activities, \eg, \textit{playing acoustic guitar} with \textit{male singing}, \textit{basketball bounce} with \textit{people crowd}, \etc.
Secondly, in Fig~\ref{fig:npmi}(b), in addition to the NPMI for consecutive occurrences of different audio-visual events, we also compute the values for the events from the same categories, which might be larger than 1.
It can be observed that the same events tend to occur repetitively in a video, especially for some events that usually happen in a short period of time, such as  \textit{people nose blowing}, \textit{people sneezing} and \textit{basketball bounce}, \etc.
Moreover, diverse consecutive dependencies also exist between different audio-visual events. 

\noindent{\bf Comparison with Existing TAL Datasets.} In Tab.~\ref{tab:tal},  we compare our UnAV-100 dataset with four popular benchmarks for temporal action localization. All these datasets are based on untrimmed videos and have relatively small scales, since 
annotating temporal boundaries for all instances in videos is labor-intensive and time-consuming.
Our UnAV-100 is the only dataset that combines both audio and visual signals to annotate instances, while others just utilize visual content in videos. Their audio tracks are usually very noisy and unrelated to the visual information, \eg, background music and narrations, thus these datasets are not suitable for joint audio-visual video understanding. Besides, these benchmarks all focus on specific domains, such as human activities, sports, cooking, \etc. By contrast, our UnAV-100 covers many different domains including human/music/sport/animal/nature, \etc, which helps machines to understand more diverse audio-visual scenes in the wild.
\newcommand{\tabincell}[2]{\begin{tabular}{@{}#1@{}}#2\end{tabular}}
\begin{table}
  \centering
  \resizebox{1.0\linewidth}{!}{
  \begin{tabular}{l|ccccc}
    \toprule
    Dataset & Videos & Classes & \tabincell{c}{Avg.\\Length} & \tabincell{c}{Avg. \\Instances} & Domains \\
    \midrule
    \midrule
    Breakfast~\cite{kuehne2014language} & 1,712 & 48 & 162s & 6 & Cooking \\
    THUMOS14~\cite{idrees2017thumos} & 413 & 20 & 212s & 15.5 & Sports \\
    ActivityNet~\cite{caba2015activitynet} & 19,994 & 200 & 115s & 1.5 & Human Activities   \\
    Charades~\cite{sigurdsson2016hollywood} & 9,848 & 157 & 30s & 6.8 & Daily Activities\\
    \midrule
    UnAV-100 (ours) & 10,790 & 100 & 42s & 2.8 & Unconstrained \\
    \bottomrule
  \end{tabular}}
  \caption{Comparison with temporal action localization datasets based on untrimmed videos.}
  \label{tab:tal}
  \vspace{-4mm}
\end{table}
\section{Implementation Details}
\noindent{\bf Feature Extraction.}
The visual features are extracted using two-stream I3D~\cite{carreira2017quo}, which inputs a set of 24 RGB and optical flow frames extracted at 25 fps.
Each frame is first resized such that the shortest side is 256 pixels, and then the center region is cropped to $224\times224$.
A 1024-d RGB or flow feature vector is obtained from the final convolutional layer of the corresponding branch of I3D.
Then, the two vectors are concatenated producing 2048-d features for each stack of 24 frames.
The audio features are extracted using VGGish~\cite{hershey2017cnn}. 
The input is a $96\times64$ log mel-scaled spectrogram extracted for each 0.96s segment, which is obtained by applying \textit{Short-Time Fourier Transform} on a 16 kHz mono audio track. Then, a 128-d feature vector can be obtained after an activation function and before a classification layer.
Here, we use 24 frames for each visual segment to temporally match with the input of the audio modality as $\frac{24}{25}=0.96$.

\noindent{\bf Network Architecture.}
In the cross-modal pyramid transformer encoder, the number of attention heads is 4 in both uni-modal and cross-modal blocks. The temporal downsampling operation is realized by using a single depth-wise 1D convolution as in~\cite{zhang2022actionformer}.
For temporal dependency modeling, the output dimension is converted as the shape of input to formulate it as a plug-and-play operation, and we just apply this operation once in our model.

\noindent{\bf Reproducibility.}
All our models are trained on a single 32GB NVIDIA Tesla V100 GPU and implemented in PyTorch deep-learning framework.
During inference, we evaluate the performances of our method on the test set of our UnAV-100 and use the best models on the validation set.    

\section{Ablation Study}
\begin{table}[t]
  \centering
  \resizebox{0.8\linewidth}{!}{
  \begin{tabular}{c|cccccc}
    \toprule
    PE & 0.5 & 0.6 & 0.7 & 0.8 & 0.9 & Avg. \\
    \midrule
    \midrule
    \rowcolor{gray!20} \checkmark & {\bf50.6} & 44.8 & {\bf39.8} & 32.4 & 21.1 & {\bf47.8} \\
     & 49.5 & {\bf45.1} & 39.7 & {\bf32.8} & {\bf21.9} & 47.0 \\
    \bottomrule
  \end{tabular}}
  \caption{Ablation study on position encoding (PE).}
  \label{tab:position}
  \vspace{-2mm}
\end{table}
\begin{table}
  \centering
  \resizebox{0.82\linewidth}{!}{
  \begin{tabular}{c|cccccc}
    \toprule
    $\lambda$ & 0.5 & 0.6 & 0.7 & 0.8 & 0.9 & Avg. \\
    \midrule
    \midrule
    0.2 & 49.9 & 45.0 & 39.6 & 32.2 & 20.7 & 46.9 \\
    0.5 & 50.1 & 45.4 & 39.8 & 32.3 & 21.2 & 47.3 \\
    \rowcolor{gray!20} 1 & {\bf50.6} & {\bf45.8} & 39.8 & 32.4 & 21.1 & {\bf47.8} \\
    2 & 49.8 & 45.3 & {\bf40.2} & {\bf33.0} & {\bf22.4} & 47.2 \\
    5 & 49.0 & 44.7 & 39.2 & 32.3 & 22.2 & 46.4 \\
    \bottomrule
  \end{tabular}}
  \caption{Ablation study on loss weight $\lambda$.}
  \label{tab:loss}
  \vspace{-2mm}
\end{table}
\begin{table}
  \centering
  \resizebox{0.85\linewidth}{!}{
  \begin{tabular}{c|cccccc}
    \toprule
    Stride & 0.5 & 0.6 & 0.7 & 0.8 & 0.9 & Avg. \\
    \midrule
    \midrule
    \rowcolor{gray!20} 8 & {\bf50.6} & {\bf44.8} & {\bf39.8} & 32.4 & 21.1 & {\bf47.8} \\
    16 & 48.9 & 44.6 & 39.0 & {\bf32.9} & {\bf21.8} & 46.7 \\
    24 & 49.7 & 44.7 & 38.5 & 31.0 & 20.9 & 47.0 \\
    \bottomrule
  \end{tabular}}
  \caption{Ablation study on temporal feature stride.}
  \label{tab:stride}
  \vspace{-2mm}
\end{table}
\begin{table}
  \centering
  \resizebox{0.86\linewidth}{!}{
  \begin{tabular}{c|cccccc}
    \toprule
    $T_{max}$ & 0.5 & 0.6 & 0.7 & 0.8 & 0.9 & Avg. \\
    \midrule
    \midrule
    192 & 49.9 & 45.2 & 39.7 & 32.6 & 21.7 & 47.0 \\
    \rowcolor{gray!20}224 & {\bf50.6} & {\bf45.8} & 39.8 & 32.4 & 21.1 & {\bf47.8} \\
    256 & 49.6 & 45.3 & {\bf39.9} & {\bf33.1} & {\bf22.3} & 47.2 \\
    \bottomrule
  \end{tabular}}
  \caption{Ablation study on maximum input sequence length.}
  \label{tab:length}
  \vspace{-2mm}
\end{table}
\begin{table}
  \centering
  \resizebox{0.95\linewidth}{!}{
  \begin{tabular}{cc|cccccc}
    \toprule
    SD & CD & 0.5 & 0.6 & 0.7 & 0.8 & 0.9 & Avg. \\
    \midrule
    \midrule
    & & 48.5 & 43.4 & 36.9 & 29.9 & 20.2 & 45.8 \\
     \checkmark & & 49.5 & {\bf45.3} & 39.7 & 32.6 & 21.2 & 46.9 \\
     & \checkmark & 49.5 & 44.8 & 39.7 & {\bf32.7} & {\bf21.7} & 46.8 \\
     \rowcolor{gray!20}\checkmark & \checkmark & {\bf50.6} & 44.8 & {\bf39.8} & 32.4 & 21.1 & {\bf47.8} \\
    \bottomrule
  \end{tabular}}
  \caption{Ablation study on dependency modeling. SD: simultaneous dependency branch; CD: consecutive dependency branch.}
  \label{tab:dependency}
  \vspace{-4mm}
\end{table}
\begin{table}[ht]
  \centering
  \resizebox{1.0\linewidth}{!}{
  \begin{tabular}{l|cccccc}
    \toprule
    Method & 0.3 & 0.4 & 0.5 & 0.6 & 0.7 & Avg. \\
    \midrule
    \midrule
    ActionFormer~\cite{zhang2022actionformer} & 73.4 & 67.5 & 57.6 & 47.6& 33.7 & 56.0 \\
    \rowcolor{gray!20}Ours & {\bf74.8} & {\bf70.1} & {\bf60.7} & {\bf48.1} & {\bf34.0} & {\bf57.5} \\
    \bottomrule
  \end{tabular}}
  \caption{Experiments on THUMOS14 dataset with only visual modality as input (mAP@[0.3:0.1:0.7] is reported).}
  \label{tab:thumos}
  \vspace{-4mm}
\end{table}

\noindent{\bf Position Encoding.}
We explore the impact of position encoding in our transformer encoder. 
As shown in Tab.~\ref{tab:position}, adding position embeddings can improve the performance by $0.8\%$ in average mAP, even though the temporal convolutions (\ie, the projection layer and downsampling operations) already leak the location information as pointed out in~\cite{xie2021segformer,zhang2022actionformer}.

\noindent{\bf Loss Weight.}
We also provide the ablation study on the loss weight $\lambda$ in our loss function.
We train the model using different loss weights $\lambda \in [0.2, 0.5, 1, 2, 5]$, and report the results in Tab.~\ref{tab:loss}.
It can be seen that the default value $\lambda=1$ can yield the best performance. 

\noindent{\bf Feature Stride.} 
In our experiments, we use stride=8 with a sliding window of 24 frames by default when extracting visual and audio features. Here, we study the performance variation using different feature strides in Tab.~\ref{tab:stride}.
Reducing the temporal feature resolution (\ie, larger strides, 16/24) leads to obvious performance degradation, which is intuitively reasonable since the model might fail to detect many short audio-visual events at a low temporal resolution.

\noindent{\bf Maximum Input Sequence Length.}
Furthermore, we vary the length of the maximum input sequences of our model, and the results are provided in Tab.~\ref{tab:length}. We can observe that our model has quite stable results when using different $T_{max}$, and $T_{max}=224$ gets the best results.

\noindent{\bf Dependency Modeling.}
Since the two branches of temporal dependency modeling aim to capture different correlations between events within a video, we run an ablation by removing each of the branches and show the results in Tab.~\ref{tab:dependency}. 
It indicates that applying each branch separately also leads to improvement, and the best result can be achieved by combing both branches to model simultaneous and consecutive dependencies at the same time.

\section{Experiments on Existing TAL Dataset}
We also conduct experiments on THUMOS14 dataset~\cite{idrees2017thumos}, a widely-used dataset for temporal action localization. The evaluation results on THUMOS14 test set using only visual input are provided in Tab.~\ref{tab:thumos}. We use the same strategy to extract features on THUMOS14 as used on UnAV-100 for both methods to keep a fair comparison.
We can see that our model outperforms ActionFormer~\cite{zhang2022actionformer} by a large margin ($+3.1\%$ mAP at tIoU=0.5), even without the cross-modal fusion strategy. 
Besides, we tried to only use the audio modality in THUMOS14 to locate actions, but got very bad results (just $4.3\%$ average mAP) on both models, which indicates that the audio tracks in THUMOS14 are quite noisy and cannot provide useful information. 

\section{More Qualitative Results}
More qualitative results are presented in Fig~\ref{fig:visual}, which includes the prediction results of our model variants using different modalities as input. Generally speaking, cross-modal perception encourages the model to obtain more correct localization results. 
For example, Fig.~\ref{fig:visual}(a) refers to the relatively constant visual information versus dramatically changing audio signals.
By integrating both modalities, the model can better judge the event boundaries. 
Besides, our audio-visual model can also get promising performance in some complex audio-visual scenarios, as in Fig.~\ref{fig:visual}(c) and Fig.~\ref{fig:visual}(d), where many audio-visual events occur concurrently or over very short periods of time. 

\section{Discussion}
\begin{figure*}[ht]
  \centering
  \setlength{\abovecaptionskip}{2mm}
    \subfloat[\label{1aa}]{
            \includegraphics[width=0.94\linewidth]{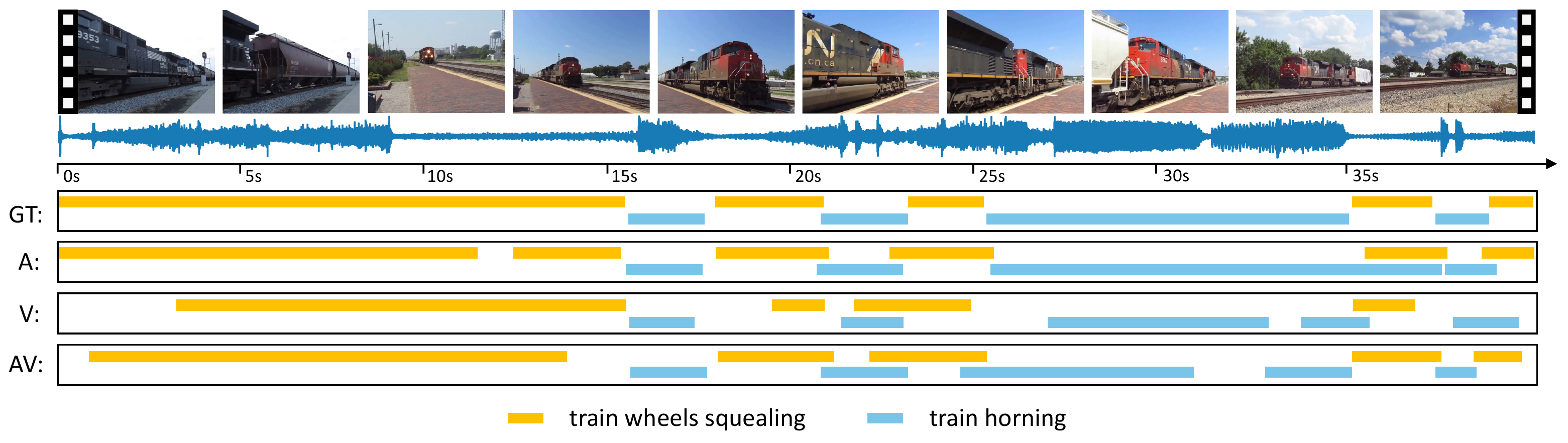}}\\
    \subfloat[\label{1bb}]{
            \includegraphics[width=0.94\linewidth]{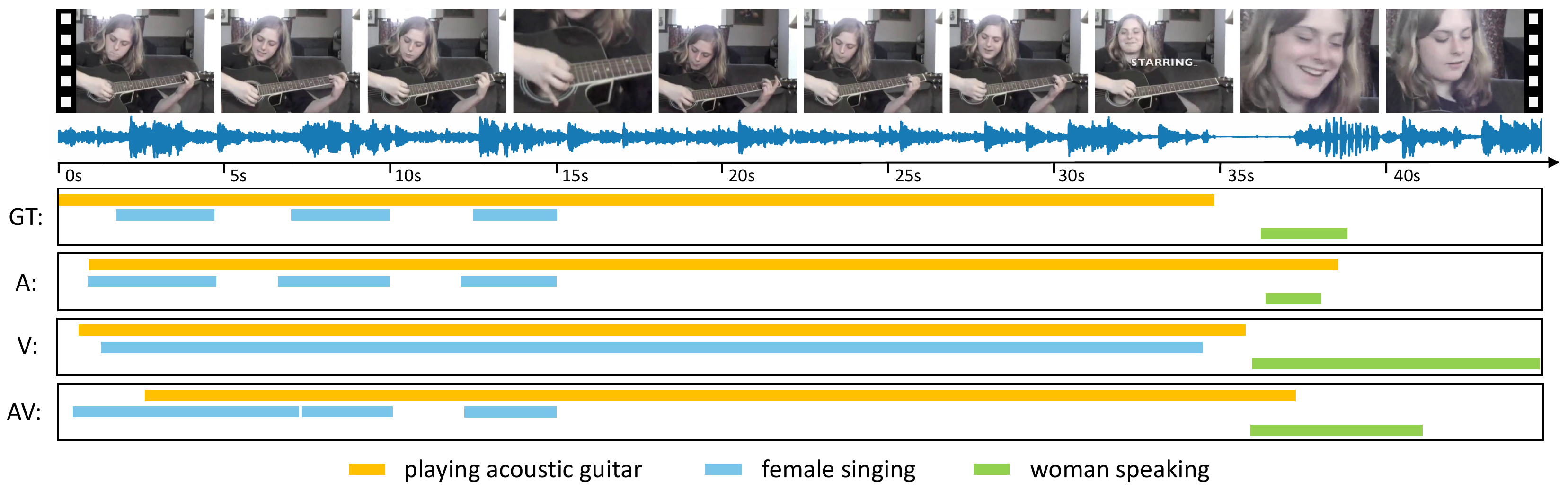}} \\
    \subfloat[\label{1cc}]{
             \includegraphics[width=0.94\linewidth]{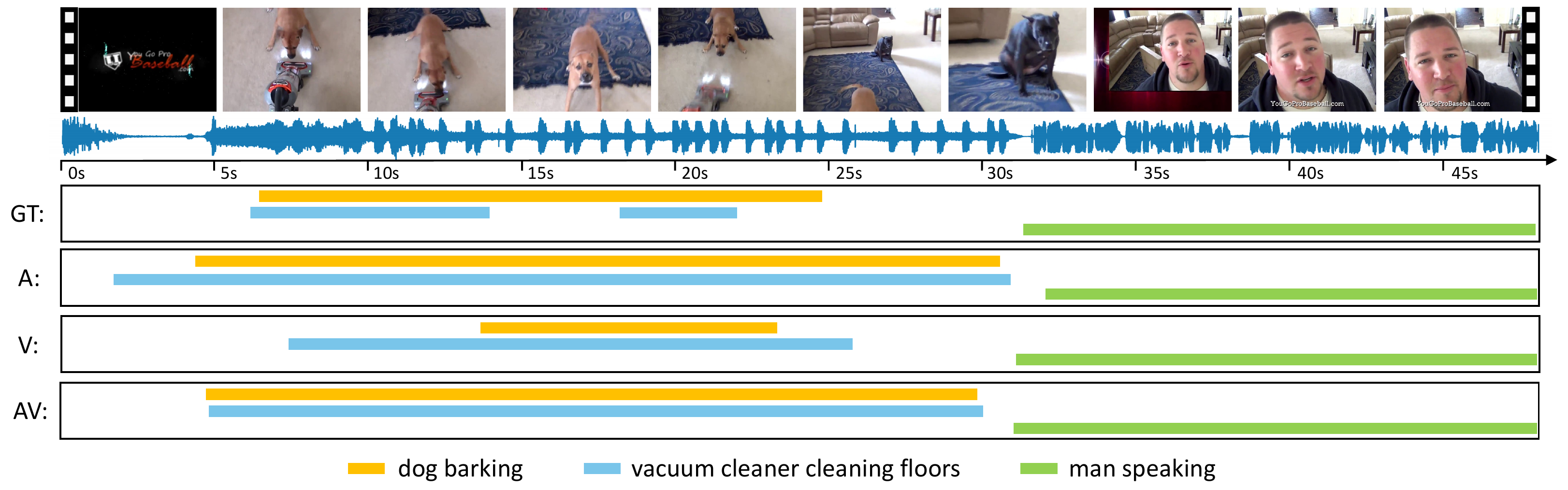}}\\
   \subfloat[\label{1dd}]{
             \includegraphics[width=0.94\linewidth]{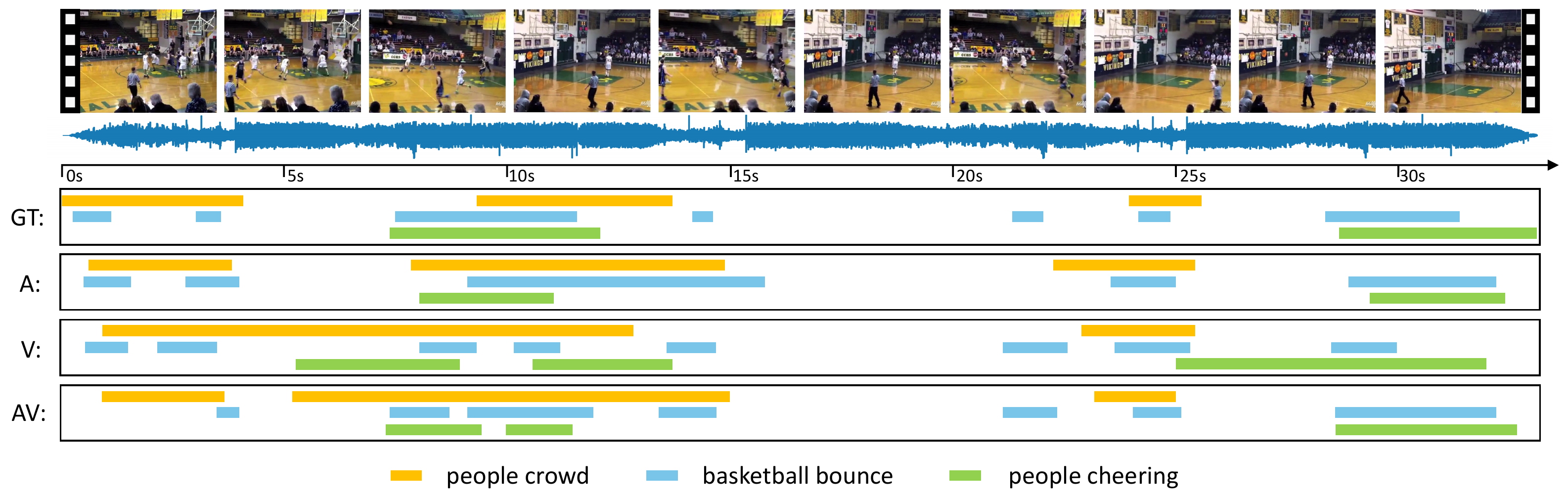}}
  \caption{More qualitative results on the UnAV-100 test set. GT: ground truth, A: the prediction of the audio-only variant, V: the prediction of the visual-only variant, AV: the prediction of our audio-visual model. We show boundaries with the highest overlap with ground truth.}
   \label{fig:visual}
\end{figure*}

\noindent{\bf Limitations.}
There is still a wide scope for exploration and improvement on the basis of our work. 
For instance, our dataset is limited to a temporal localization task. 
We will explore other audio-visual problems, such as representation learning and sound source localization in real-life and complex scenarios in our subsequent study.
Besides, although our model can obtain a promising performance, as a baseline, its capability is still limited in some complex situations.
For example, in Fig.~\ref{fig:visual}(c), the model gets an incorrect boundary of the \textit{dog barking} event when the barking brown dog is out of the screen and a non-barking black one can be seen. This indicates that our model might fail to effectively filter out interference information for such a difficult case.
And the model might also fail to predict precise boundaries when one modality persists while another disappears for a short period of time (\eg, the event of \textit{vacuum cleaner cleaning floors} in Fig.~\ref{fig:visual}(c)).
In addition, for some instant events with very short duration (\eg, \textit{basketball bounce} in Fig.~\ref{fig:visual}(d)), our model might get unsatisfactory results.
Overall, dense-localizing audio-visual events is inherently a very challenging task, and it requires the model to have a strong fine-grained cross-modal understanding ability.
Therefore, more advanced models that could better solve the above difficulties are expected to boost performance further.
We hope our work as the first attempt at untrimmed audio-visual video understanding can inspire more people to explore the field.

\noindent{\bf Ethic concerns and biases.}
Our UnAV-100 is sourced from VGGSound dataset~\cite{chen2020vggsound} that has already tried to mitigate ethical issues. During data collection, we made further efforts to manually check all videos to avoid mature, sensitive, or offensive content. 
Besides, our UnAV-100 follows the natural distribution of instances present on the website, which may reflect some biases in topics. For example, there are more {\textit{man/woman speaking}} events than other categories. Efforts have been made to mitigate such imbalance.


\end{document}